\documentclass[10pt, conference, compsocconf]{IEEEtran}
\IEEEoverridecommandlockouts
\usepackage{CJKutf8}

\usepackage[section]{placeins}

\usepackage{makecell}

\usepackage{booktabs}

\ifCLASSINFOpdf
   \usepackage[pdftex]{graphicx}
   \graphicspath{{..}{../jpeg/}}
   \DeclareGraphicsExtensions{.pdf,.jpeg,.png}
\else
   \usepackage[dvips]{graphicx}
   \graphicspath{{../eps/}}
   \DeclareGraphicsExtensions{.eps}
\fi
\makeatletter
\newcommand{\linebreakand}{%
  \end{@IEEEauthorhalign}
  \hfill\mbox{}\par
  \mbox{}\hfill\begin{@IEEEauthorhalign}
}
\makeatother
\usepackage{amsmath}

\begin{document}
\begin{CJK*}{UTF8}{gbsn}
\begin{CJK*}{UTF8}{bsmi}
%
\title{SPSQL: Step-by-step Parsing Based Framework for Text-to-SQL Generation}


\author{\IEEEauthorblockN{Ran Shen}
\IEEEauthorblockN{Marketing Service Center\\
State Grid Zhejiang Electric\\
Power Co., Ltd\\
Hangzhou, China\\
shen\_ran@zj.sgcc.com.cn}
\and
\IEEEauthorblockN{Gang Sun}
\IEEEauthorblockN{Marketing Service Center\\
State Grid Zhejiang Electric\\
Power Co., Ltd\\
Hangzhou, China\\
sun\_gang@zj.sgcc.com.cn}
\and
\IEEEauthorblockN{Hao Shen}
\IEEEauthorblockN{Marketing Department\\
State Grid Zhejiang Electric\\
Power Co., Ltd\\
Hangzhou, China\\
shen\_hao@zj.sgcc.com.cn}
\linebreakand
\IEEEauthorblockN{Yiling Li}
\IEEEauthorblockN{Marketing Service Center\\
State Grid Zhejiang Electric\\
Power Co., Ltd\\
Hangzhou, China\\
li\_yiling@zj.sgcc.com.cn}
\and
\IEEEauthorblockN{Liangfeng Jin}
\IEEEauthorblockN{Marketing Service Center\\
State Grid Zhejiang Electric\\
Power Co., Ltd\\
Hangzhou, China\\
jin\_liangfeng@zj.sgcc.com.cn}
\and
\IEEEauthorblockN{Han Jiang\IEEEauthorrefmark{1}
\thanks{\IEEEauthorrefmark{1}Corresponding Author}}
\IEEEauthorblockN{College of Computer Science\\
and Technology\\
Zhejiang University\\
Hangzhou, China\\
jianghan97@zju.edu.cn}
}


%


\maketitle

\begin{abstract}
Converting text into the structured query language (Text2SQL) is a research hotspot in the field of natural language processing (NLP), which has broad application prospects. In the era of big data, the use of databases has penetrated all walks of life, in which the collected data is large in scale, diverse in variety, and wide in scope, making the data query cumbersome and inefficient, and putting forward higher requirements for the Text2SQL model. In practical applications, the current mainstream end-to-end Text2SQL model is not only difficult to build due to its complex structure and high requirements for training data, but also difficult to adjust due to massive parameters. In addition, the accuracy of the model is hard to achieve the desired result. Based on this, this paper proposes a pipelined Text2SQL method: SPSQL. This method disassembles the Text2SQL task into four subtasks——table selection, column selection, SQL generation, and value filling, which can be converted into a text classification problem, a sequence labeling problem, and two text generation problems, respectively. Then, we construct data formats of different subtasks based on existing data and improve the accuracy of the overall model by improving the accuracy of each submodel. We also use the named entity recognition module and data augmentation to optimize the overall model. We construct the dataset based on the marketing business data of the State Grid Corporation of China. Experiments demonstrate our proposed method achieves the best performance compared with the end-to-end method and other pipeline methods.

\end{abstract}


%
\IEEEpeerreviewmaketitle

\section{Introduction}
The database stores massive amounts of high-value data. Taking the State Grid Corporation of China as an example, the data it stores and manages has reached the petabyte level \cite{2022Intelligent,2012Big}. The acquisition and analysis of these data require interactive operation with the database by writing structured query language (SQL) queries, which brings inconvenience to ordinary users. Manually written SQL query statements are prone to errors when there are complex query conditions, which leads to insufficient data mining depth and weak data cashability. It is urgent to realize the transformation of human-computer interaction mode through artificial intelligence (AI) technology to improve the efficiency of data analysis and mining.

Text-to-SQL (Text2SQL) is an AI technology that converts natural language into SQL. The Text2SQL model generates SQL statements by understanding the question, analyzing the table content and schema of the database. The application of Text2SQL makes it unnecessary for users to study the specific structure and query syntax of the database \cite{2020Database}. It also reduces the threshold of data analysis and improves the utilization of data and the efficiency of data querying.

Text2SQL has pipeline methods and deep learning methods. The pipeline method converts the natural language into an intermediate expression through the model and then maps the intermediate expression into the corresponding SQL query statements. NChiql \cite{2001Method}, NaLIR \cite{2014Constructing}, and TEMPLAR \cite{2019Bridging} are representative pipeline methods. Statistical parser \cite{2004Modern} and SPARQL \cite{0Template} are classic “intermediate expressions”. The pipeline method disassembles the whole model into individual steps, which makes it easy to achieve high accuracy in practical applications. However, this method relies on the regular description of natural language queries, making it unable to handle complex and changeable natural language descriptions \cite{2020Research}. The deep learning method uses an end-to-end neural network to complete the conversion of natural language to SQL. Early deep learning methods mainly include the Seq2Seq$+$Attention model \cite{dong2016language}, Seq2Seq$+$Copying model \cite{wang2018pointing}, Seq2SQL \cite{2017Seq2SQL}, SQLNet \cite{2017SQLNet}, etc. Later, methods are mainly based on pre-trained models, such as RoBERTa \cite{liu2019roberta}, XLNet \cite{yang2019xlnet}, ERNIE \cite{zhang2019ernie}, etc. The deep learning method can not only better capture the complex semantics of query sentences, but also no longer extract features manually. However, its end-to-end architecture leads to complex models, low accuracy in practical applications, and weak practicability in industry.

In the face of the practical application of Text2SQL, we abandon the end-to-end architecture of deep learning methods, which is difficult to train and adjust parameters to adapt to specific tasks due to the complicated and cumbersome structure. Instead, we propose a Text2SQL framework with pipeline structure to disassemble the Text2SQL task into four simple subtasks: table selection, column selection, SQL generation, and value filling, thereby transforming the original complex Text2SQL model into four flexible and lightweight submodels: text classification model, sequence labeling model, and two text generation models. In this way, improving the accuracy of the whole model can be changed into improving the accuracy of each submodel, so that different optimization methods can be adopted according to the characteristics of each submodel. Previous pipeline methods such as SPARQL, NaLIR, etc., rely on templates and manually designed features \cite{2018Research}, resulting in inflexibility and poor model mobility \cite{2020Research}. Therefore, in terms of models, we convert each subtask into a standard natural language processing (NLP) deep learning model, which frees a large number of labor costs from rule writing and feature design in a data-driven way, to improve the flexibility and automation of each subtask; In addition, since the pre-trained model can be fine-tuned directly according to the needs of downstream tasks, which eliminates the process of starting from scratch and saves the time and resources of model training, we adopt the “pre-trained model plus fine-tuning” approach for each submodel. In terms of data, we augment data in three ways to meet the requirements of the model: 1. randomly replace keywords in the question, 2. randomly add, delete or replace column names in the text, 3. introduce simBERT for similar text generation. Besides, we introduce a named entity recognition (NER) module in prediction. It associates and replaces the named entity in the input question with that in the database, and then replaces it back after the model outputs the SQL statement. It can solve the problem of generating wrong SQL statements when the training data is too little to accurately identify entities.

Overall, the main contributions of this paper are as follows: 

1. The accuracy of the large model of the end-to-end architecture can only reach about 70\% \cite{lan2019albert}, which cannot meet the needs of practical applications, and the large model cannot be targeted for specific problems in the application process. Therefore, we decompose Text2SQL into four simple subtasks: table selection, column selection, SQL generation, and value filling. Then, we construct a text classification submodel, a sequence labeling submodel, and two text generation submodels according to the characteristics of the task, which are all pre-trained and fine-tuned. In addition, we adopt different pre-training models and data construction methods according to the characteristics of the model.

2. The data quality provided in the actual application scenario is difficult to meet the requirements of the model, so a large amount of manual construction cost is required in this case. Therefore, we propose three methods for data augmentation to improve the performance of the model: random replacement of keywords in the question, random addition, deletion, or replacement of column names in the text, and introduction of simBERT for similar text generation.

3. The insensitivity of the model to entity information in natural language questions can lead to typos and omissions of named entities in the generated SQL statements. Therefore, we creatively introduce the NER module. The module replaces the named entity in the question with the one existing in the database before the question is input into the model. After the model outputs the SQL statement, the module replaces the named entity back. In this way, the accuracy of SQL statement generation can be improved.

The rest of this paper is organized as follows: In section \uppercase\expandafter{\romannumeral2}, a brief review of related work with Text2SQL and the pre-trained language model will be given. Section \uppercase\expandafter{\romannumeral3} presents our proposed work. Section \uppercase\expandafter{\romannumeral4} presents and analyzes the experimental results. Finally, the conclusion is illustrated in Section \uppercase\expandafter{\romannumeral5}.
\section{RELATED WORK}
\subsection{Text2SQL}
The pipeline method is the traditional method of Text2SQL, featuring “intermediate expressions” and “rules”. In 2004, Popescu et al. \cite{2004Modern} used a statistical parser as a “plug-in” to correctly map parsed questions to corresponding SQL queries. In 2012, Unger et al. \cite{0Template} proposed a method based on SPARQL syntax and WordNet. In 2014, Li et al. \cite{2014Constructing} proposed a NaLIR system using a “parse tree” for intermediate expression \cite{2022Technical}. In 2019, Baik et al. \cite{2019Bridging} proposed the TEMPLAR system, which enhances the existing pipeline-based natural language database interface with SQL query log information. The pipelined method is relatively simple in structure and practical in industry, but it cannot solve the problem of information loss during conversion due to the existence of intermediate expression. Moreover, in the face of different tasks, a lot of rules need to be written manually, so the degree of automation is low. At this time, the Text2SQL method based on deep learning came into being. In 2017, Zhong et al. \cite{2017Seq2SQL} proposed Seq2SQL, which divides slots according to different components of SQL query statements, and then generates SQL query statements based on slot filling \cite{2020Database}. In the same year, Xu et al. \cite{2017SQLNet} proposed SQLNet based on Seq2SQL, which greatly reduced the syntax errors of SQL query statements by filling templates with predefined SQL query statements, thus improving accuracy. Since 2018, with the development of pre-trained language models such as ELMo \cite{2018Deep}, GPT \cite{radford2018improving}, BERT \cite{devlin2018bert}, and T5 \cite{raffel2020exploring}, more and more researchers have applied them to Text2SQL, resulting in methods based on pre-trained language models such as RoBERTa, XLNet, ERNIE, FastBERT \cite{liu2020fastbert}, MobileBERT \cite{sun2020mobilebert}, and MTL-BERT \cite{2021MTL-BERT}. The deep learning method is highly automated, but the interpretability of the model is worse than that of the pipeline method, and it usually requires a huge structure to support complex tasks.
\subsection{The Pre-trained Language Model}
Pre-training technology refers to pre-designing the network structure and inputting the encoded data into the network for training to increase the generalization ability of the model \cite{2020Resea}. The above network is called the pre-trained model. Pre-trained models include static pre-trained models and dynamic pre-trained models. In 2003, Bengio et al. \cite{bengio2000neural} proposed the NNLM model, which gave birth to a series of word vector methods such as Word2Vec \cite{mikolov2013efficient}, Glove \cite{pennington2014glove}, FastTest \cite{joulin2016bag}, etc. These models belong to static pre-trained models, which are difficult to solve the problem of polysemy and have limited improvement for downstream tasks. After that, the dynamic pre-trained model was proposed. In 2018, ELMo was proposed \cite{2018Deep}. ELMo uses a bidirectional long short-term memory (LSTM) artificial neural network for pre-training, which can effectively deal with polysemy. In the same year, GPT combined unsupervised pre-training with supervised fine-tuning for the first time, more suitable for downstream tasks \cite{radford2018improving}. However, as a unidirectional language model, GPT has a limited ability to model semantic information \cite{2020Overview}. Later, BERT first used the bidirectional Transformer \cite{vaswani2017attention}
in a language mode, solving the problem of GPT models discarding the next text to prevent information leakage \cite{2020Resea}. The emergence of BERT has opened a new era of pre-training technology. Since then, lots of pre-training language models have emerged, such as BERT-based improved models RoBERTa and ALBERT \cite{albert}, the autoregressive language model XLNET, the general unified framework T5, and so on.

Among the above models, ELMo is more suitable for short text tasks, XLNET is more suitable for long text tasks, GPT's understanding of context is worse than BERT, and T5 has more parameters than BERT. Therefore, we choose BERT, which is not critical of text length and performs well in text classification and sequence labeling, as the pre-training model in the table selection subtask and column selection subtask. Limited by the model framework, BERT is not ideal for text generation tasks. Compared with BERT, T5 is also universal and has a Chinese pre-trained model, which is only slightly inferior to BERT in terms of parameter quantity. Therefore, we choose T5 as the pre-trained model of the text generation subtask.

\section{PROPOSED WORK}
\subsection{Model Overview}
The Text2SQL task aims to convert natural language into structured query language. Its formal definition is as follows:
\begin{equation}
    Q \xrightarrow{S_s,S_c} R.
\end{equation}
Specifically, natural language imperative sentences or interrogative sentences that can indicate query intent are expressed as
\begin{equation}
    Q=\{w_1,w_2,...,w_n\},
\end{equation}
where $w_n$ represents the $n$th word that makes up a sentence. The database schema is expressed as
\begin{equation}
\begin{aligned}
    S_s=\{s_1,s_2,&...,s_i,...,s_m\},\\
    s_i=\{(t_i,c_{i1}),(t_i,c_{i2}),&...,(t_i,c_{ij}),...,(t_i,c_{il})\},
\end{aligned}
\end{equation}
where $m$ represents the number of tables in the database, $s_i$
represents the structure of the $i$th table in the database, $t_i$ represents the name of the $i$th table, $c_{ij}$ represents the name of the $j$th column in the $i$th table, and $l$ represents the number of columns in the $i$th table. The database content is expressed as
\begin{equation}
\begin{aligned}
    S_c=\{T_1,T_2,&...,T_i,...,T_m\},\\
    T_i=\{C_{i1},C_{i2},&...,C_{ij},...,C_{il}\},\\
    C_{ij}=\{C&_{ij}^1,C_{ij}^2\},
\end{aligned}
\end{equation}
where $T_i$ represents the $i$th table in the database, $C_{ij}$ represents the $j$th column in the $i$th table, $C_{ij}^1$ represents the content of the $j$th column in the $i$th table, and $C_{ij}^2$ represents the type of the $j$th column in the $i$th table (such as time, numerical value, text, etc.). The corresponding SQL query sentence of the natural language query sentence $Q$ is represented as $R$.

\begin{figure}
  \centering
  \includegraphics[width=2.5in]{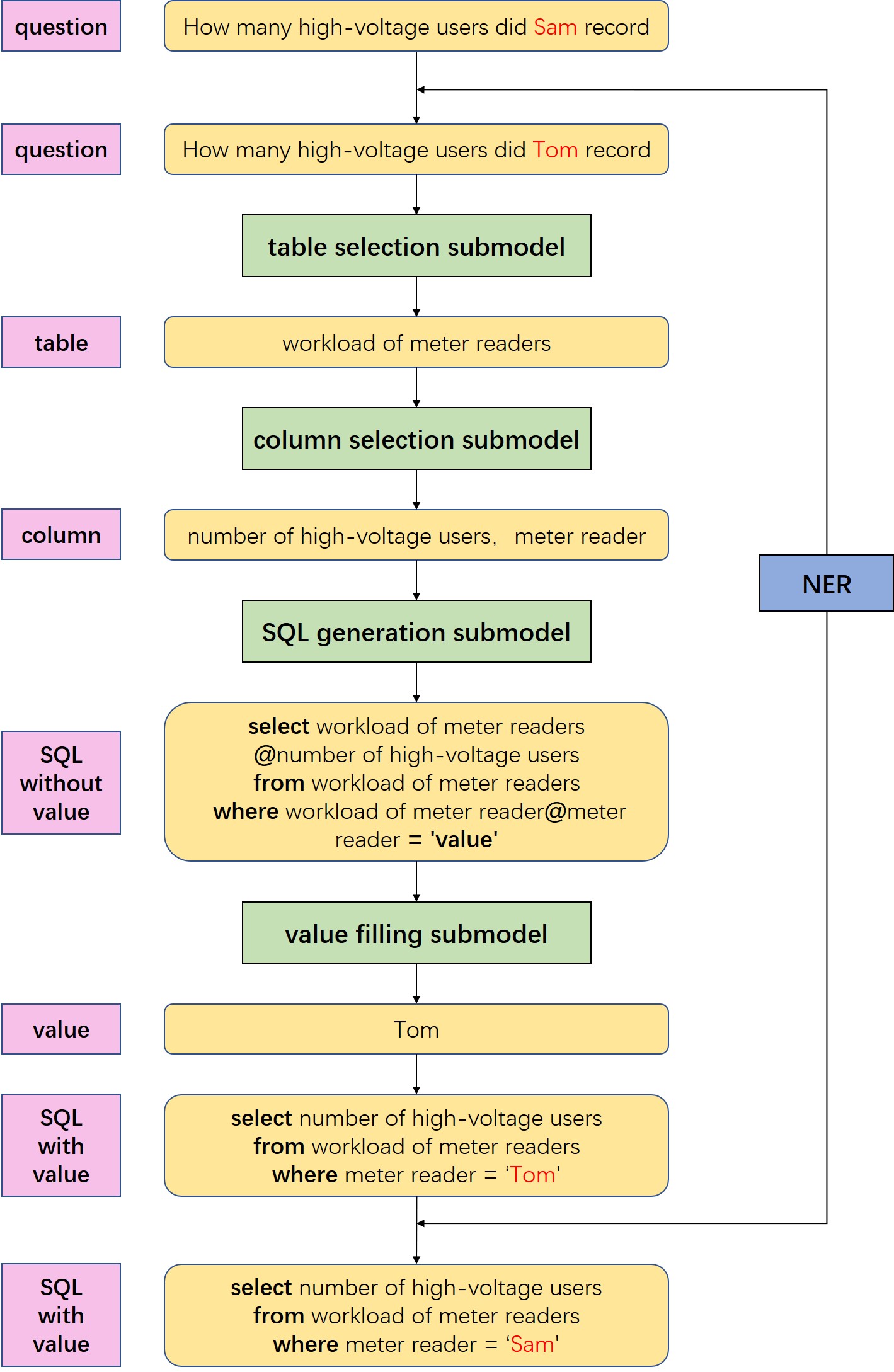}
  \caption{Frame diagram of our method.}\label{fig1}
\end{figure}

The framework of our method is shown in Figure \ref{fig1}. In this paper, the trained text classification model, sequence labeling model, and two text generation models are used to construct the pipeline structure, which receives the question input by a user, and generates the corresponding SQL query statements through the table selection task, column selection task, SQL generation task and value filling task. Specifically, in the first step, the input question is matched to the corresponding table in the database through the table selection model. Its formal definition is as follows:
\begin{equation}
    Q \xrightarrow{S_s} t.
\end{equation}
In the second step, the selected table is combined with the question, and the corresponding columns in the database are matched by the column selection model. Its formal definition is as follows:
\begin{equation}
    (Q,t) \xrightarrow{S_s} c.
\end{equation}
In the third step, the tables and columns selected in the first two steps are combined with the question to generate the SQL query statement without values through the SQL generation model. Its formal definition is as follows:
\begin{equation}
    (Q,t,c) \xrightarrow{S_s} R_s,
\end{equation}
where $R_s$ represents the SQL query statement without values. The fourth step is to combine the SQL query statement generated in the previous step with the question, input it into the text generation model for value filling, output the corresponding value, and integrate it to obtain the standard SQL query statement. Its formal definition is as follows:
\begin{equation}
\begin{aligned}
    (Q,R_s) \xrightarrow{S_s,S_c} R_v,\\
    (R_v,R_s) \xrightarrow{} R,
\end{aligned}
\end{equation}
where $R_v$ represents values in the SQL query statement.
\subsection{Optimization measures for submodels}
The pipeline structure facilitates our targeted optimization of the four subtasks to improve the overall accuracy. We adopt different model optimization and data construction methods according to the characteristics of each subtask.
\subsubsection{Selection and optimization of submodels}
In the table selection subtask and the column selection subtask, we select BERT as the pre-trained model due to its good performance in the text classification task and the sequence labeling task, and construct the table selection submodel and column selection submodel in the way of BERT$+$finetune. The generation of SQL statements is the key and aporia in Text2SQL tasks. Even BERT, which is known for its versatility, has hit the wall in the generation task. To add more input information to the model to facilitate SQL generation, we divide the SQL generation task into two stages: Generate the SQL statement without values, and then generate values to fill in the corresponding positions to form a complete SQL statement. We convert these two stages into SQL (without value) generation model and value filling model respectively, and build them in the way of T5$+$finetune. In addition, to reduce the learning difficulty of the SQL generation model, we use the copy mechanism \cite{vinyals2015pointer}, which can directly select the required fields from the input text for direct generation, and use the beam-search method to avoid generating illegal SQL query statements.
\subsubsection{Data construction of submodels}
We parse the table selection task into a binary classification problem, that is, whether the table selected from the database is mentioned in the question. Therefore, we hope to input a question and a table name in the database to output a yes or no answer through the model. Its formal definition is as follows:
\begin{equation}
\begin{aligned}
   (Q,t_i) \xrightarrow {}o,\\
o=\left\{
\begin{aligned}
		1 \quad \mbox{if} ~ t_i  \in Q     \\
		0 \quad \mbox{if} ~ t_i  \notin Q,
\end{aligned}
\right.
\end{aligned}
\end{equation}
where the output result is expressed as $o$, when the table is mentioned in the question, the value is 1, otherwise, the value is 0. We construct the data format accordingly: \{“input”: “question extra0 table\_name”, “label”: 0 or 1\}, where extra0 is the separator. Then, based on the question-SQL pair dataset, we construct a dataset for the binary classification model through this data format.

The column selection task requires the model to label the columns mentioned in the question. Therefore, we hope to input a question, the table name mentioned in the question, and the names and types of all columns in the table (Column types are inputted to enrich the input to the model.) to output information about whether columns hit or not through the model. Its formal definition is as follows:
\begin{equation}
\begin{aligned}
    (Q,s_i,C_i^2) \xrightarrow {}O,\\
    O=\{o_1,o_2,...,o_a\},
    \end{aligned}
\end{equation}
where the output sequence is expressed as $o$, $a$ represents the sequence length. The output sequence is as long as the input sequence, and the positions correspond to each other. We use column separators to mark column hit information. When a column hit occurs, the label at the position of its column separator is marked as ‘B-C’; if it does not hit, it is marked as ‘B-N’; the labels at other positions are all marked as ‘O’. We construct the input data format accordingly: \{“input”: “question extra0 table\_name extra1 name\_of\_column\_1 type\_of\_column\_1 ... extra1 name\_of\_column\_n type\_of\_column\_n”\}, where “extra0” is the table separator and “extra1” is the column separator. Then, based on the question-SQL pair dataset, we construct a dataset for the sequence labeling model through this data format.

For the SQL generation model, we hope to input a question, table names, and column names mentioned in the question to output SQL statements without values through the model. Its formal definition is as follows:
\begin{equation}
    (Q,t,c) \xrightarrow {}R_s.
\end{equation}
In terms of data construction, we replace specific table and column names with corresponding identifiers to generate SQL statements without values. Besides, to make the model better understand the relationship among questions, tables, and columns during training, we repeat the question at the end of the input data. Specifically, the format is: \{“input”: “question extra50 extra54 name\_of\_table\_1 extra51 extra0 name\_of\_column\_1 ... extra51 extra(n-1) name\_of\_column\_n…extra50 extra(53+m) name\_of\_table\_m ... extra53 question”\}, where extra50 is the table separator, extra51 is the column separator，extra53 is the question separator，extra54 is the identifier corresponding to the following table, extra0 is the identifier corresponding to the following column. The output format is: \{“SQL”: the SQL statements without values\}. Then, based on the question-SQL pair dataset, we construct a dataset for the SQL generation model through this data format.

In the value filling model, we input the question and the SQL statement generated in the previous stage into the model, hoping to get the specific value. Its formal definition is as follows:
\begin{equation}
    (Q,R_s) \xrightarrow {}R_v.
\end{equation}
We construct the input data format accordingly: \{“input”: “question [SEP] select name\_of\_table\_1 @ name\_of\_column\_1 from name\_of\_table\_1 where name\_of\_table\_1 @ name\_of\_column\_1 = ‘extra1’ and name\_of\_table\_1 @ name\_of\_column\_3 = ‘extra2’ [SEP] question”, “value”: “extra1 value\_1 extra2 value\_2”\}. The question and the SQL statement are separated by a special separator “[SEP]”, extra1 and extra2 represent identifiers of values. Then, based on the question-SQL pair dataset and the SQL statement generated in the previous stage, we construct a dataset for the SQL generation model through this data format.

Datasets formed by the data formats mentioned above are conducive to improving the training efficiency of each subtask.
\subsection{Optimization measures for practical application scenarios}
Apart from the above-mentioned optimization measures for submodels, we consider practical applications and propose optimization measures for the entire model in terms of data augmentation and NER.
\subsubsection{data augmentation}
\begin{figure}
  \centering
  \includegraphics[width=3in]{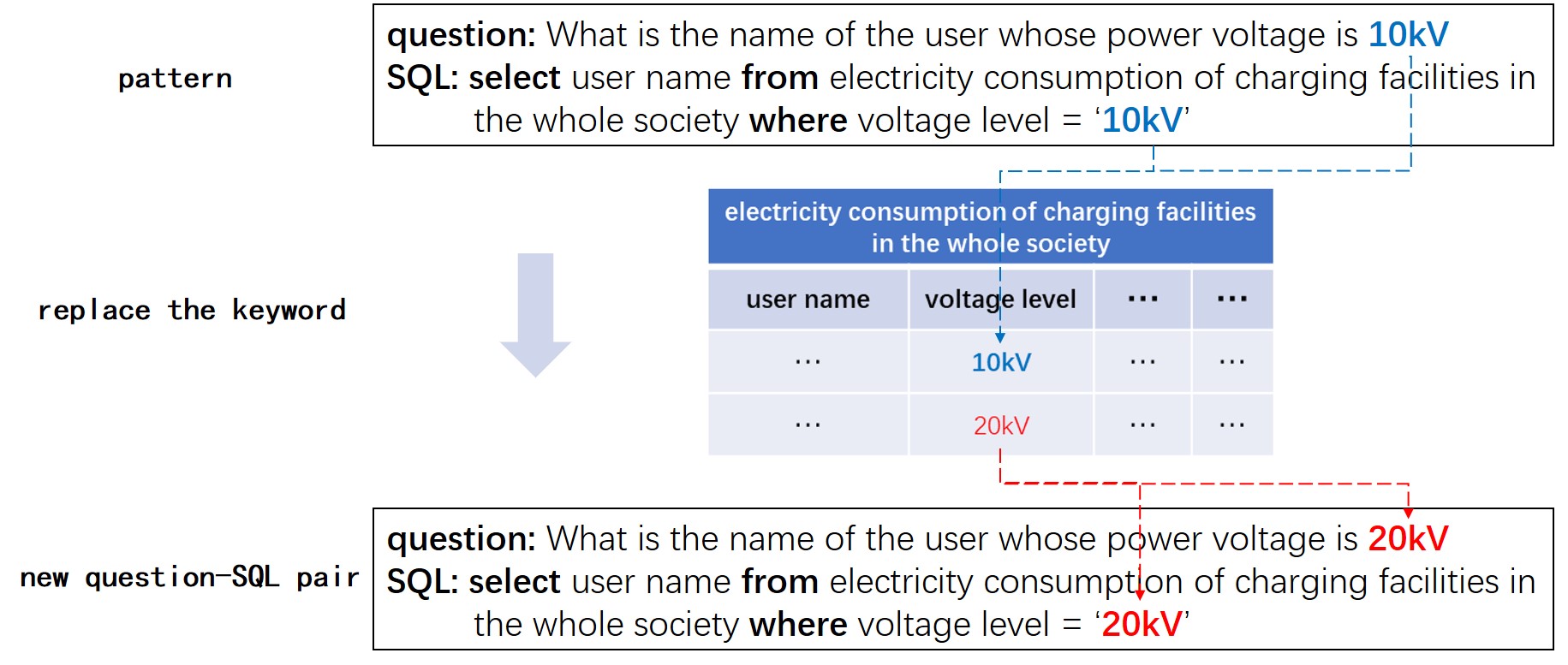}
  \caption{Replace the keyword in the question.}\label{fig2}
\end{figure}

The data generated in the real scene has the characteristics of poor interpretability and scattered distribution. Some data are difficult to obtain due to the personal information and trade secrets involved, or need to take effort to process private information. Therefore, both in quality and quantity, these data are difficult to meet the demands of Text2SQL tasks, requiring a lot of labor costs in data cleaning and data construction. Therefore, we use three ways to augment data:

1. Replace keywords in the question. We replace values of the database that exist in the question. As shown in Figure \ref{fig2}.

\begin{figure}
  \centering
  \includegraphics[width=3in]{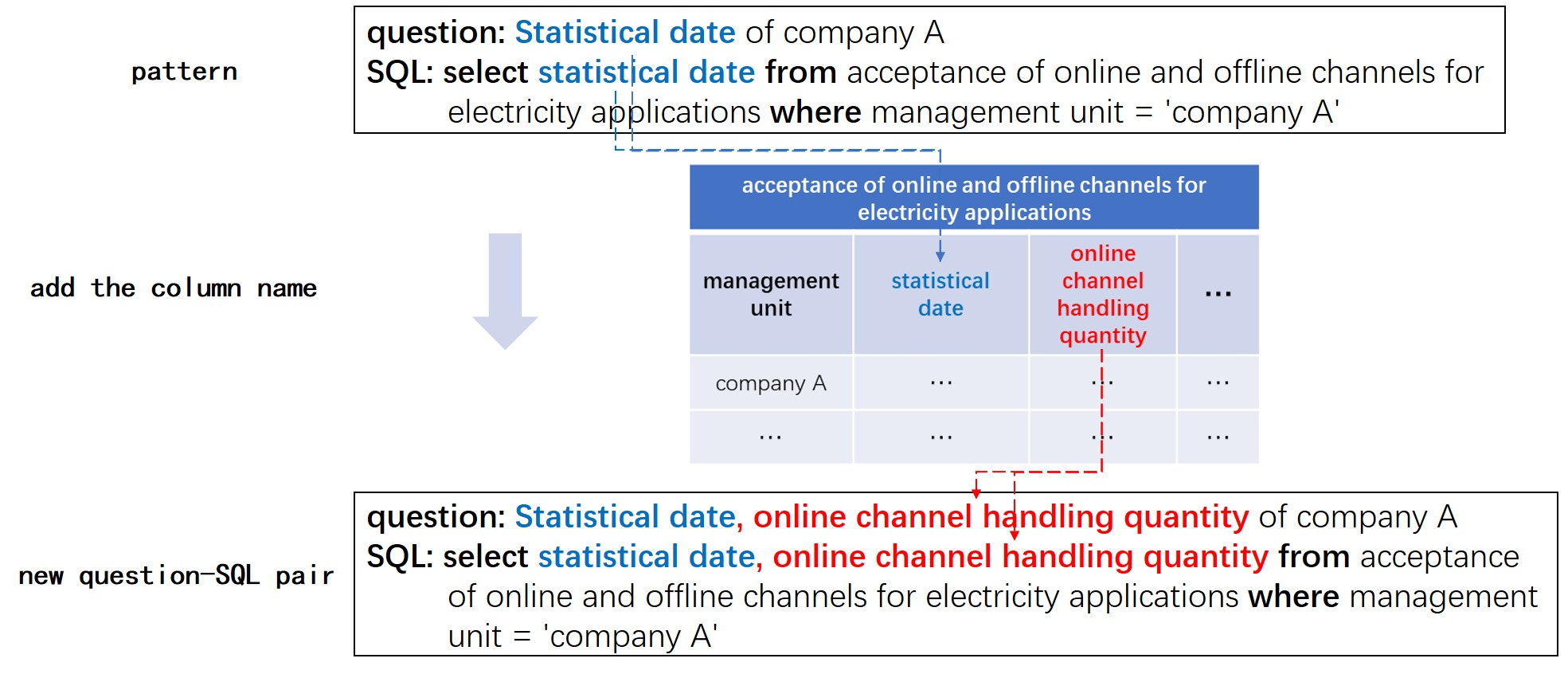}
  \caption{Replace the column name in the text.}\label{fig3}
\end{figure}

2. Add, delete, or replace column names in the text. We replace column names of the database that exist in the input. As shown in Figure \ref{fig3}. 

\begin{figure}
  \centering
  \includegraphics[width=2.5in]{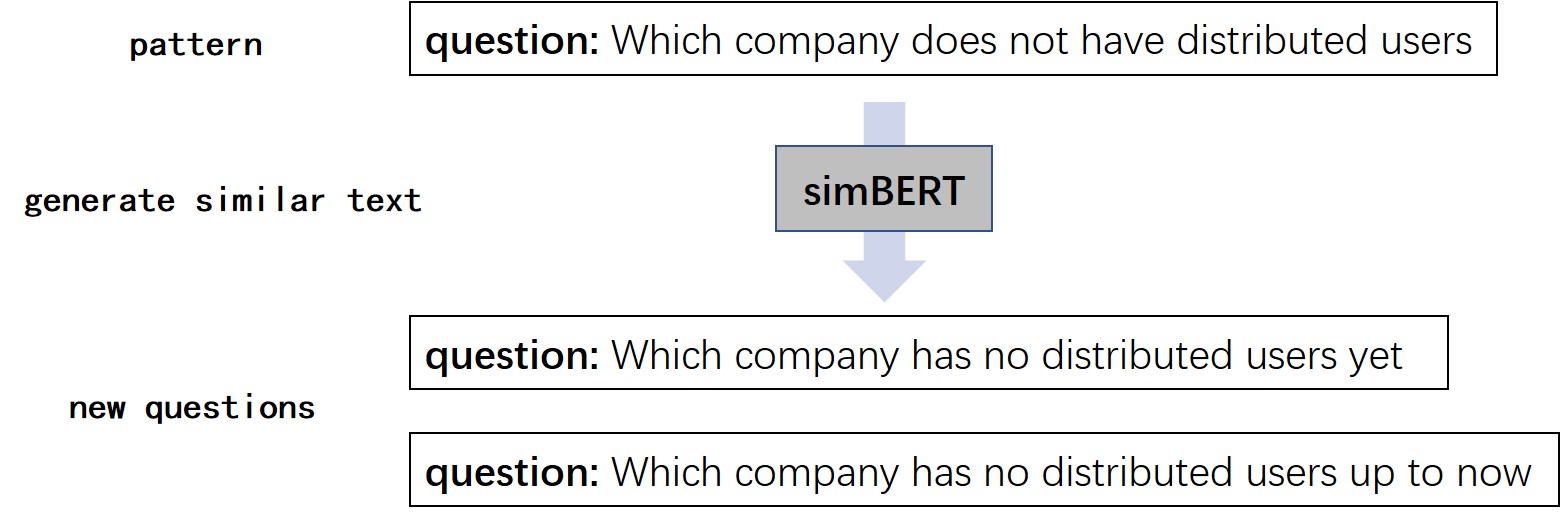}
  \caption{Schematic diagram of simBERT generating similar questions.}\label{fig4}
\end{figure}

3. Introduce simBERT for similar text generation. We input the question into simBERT to generate multiple texts that are semantically similar to the question. As shown in Figure \ref{fig4}.
\subsubsection{NER}
\begin{figure}
  \centering
  \includegraphics[width=3in]{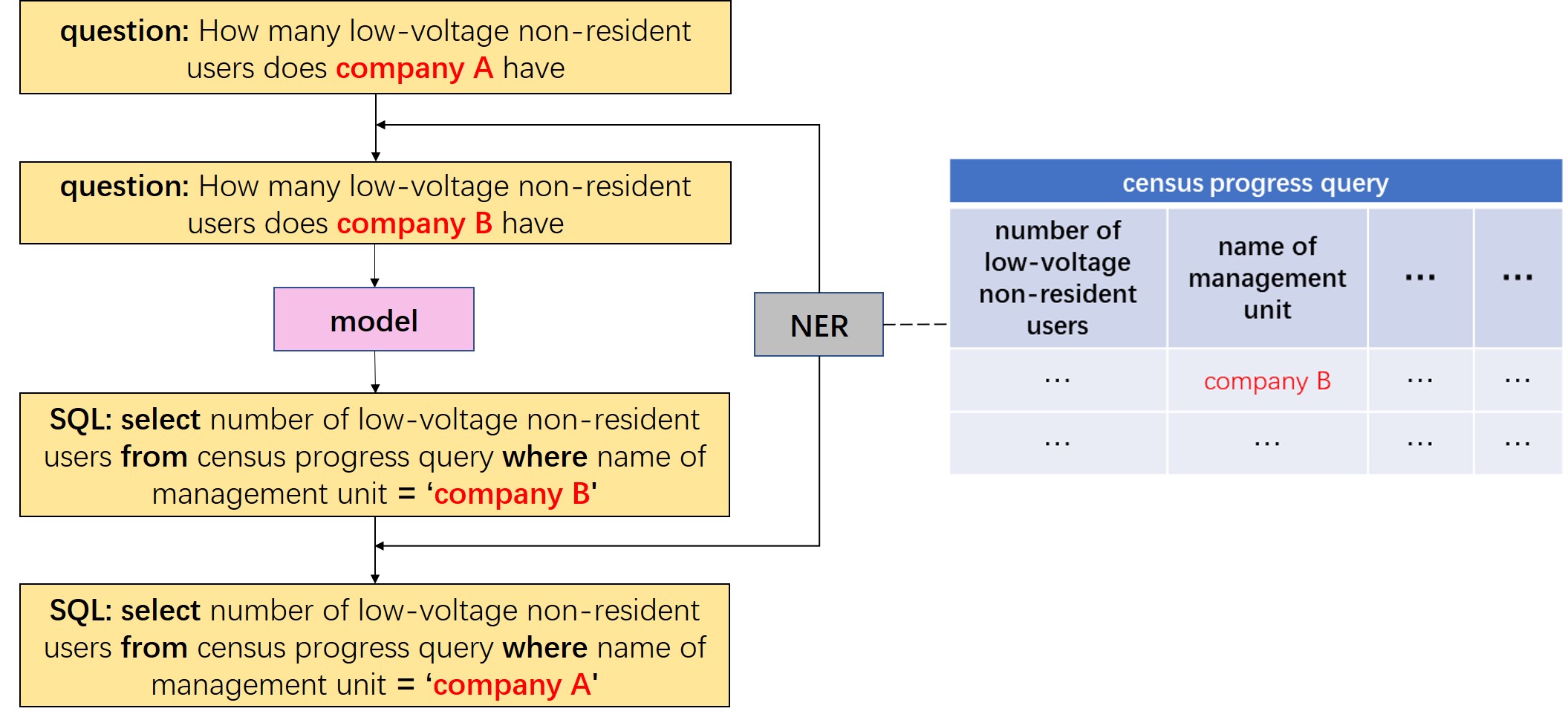}
  \caption{Frame diagram of our method.}\label{fig5}
\end{figure}

In practical applications, the variety of entities in natural language questions brings great difficulties to model recognition, which leads to typos and omissions of named entities in the generated SQL queries. Therefore, as shown in Figure \ref{fig5}, we creatively constructed a NER module. This module identifies the named entities in the input text, then associates them with the database, and then replaces them with the representations of the named entities that exist in the database. The replaced problem input model generates SQL statements according to the normal process. After the model outputs the SQL statement, NER replaces the replaced named entity to ensure that the SQL statement corresponds to the problem. The involvement of NER reduces the difficulty of model identification, thus improving the accuracy of the overall model.
\section{EXPERIMENT}
\subsection{Datasets}
\begin{figure}
  \centering
  \includegraphics[width=2.5in]{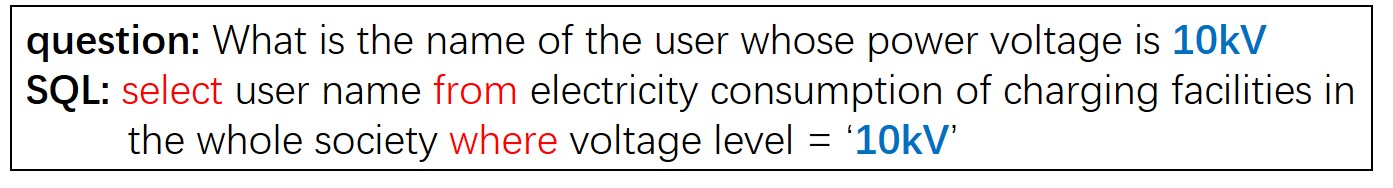}
  \caption{The specific format of each sample.}\label{fig6}
\end{figure}

We construct datasets based on data from the marketing scenario of State Grid Corporation of China. Since lots of data involve personal privacy, the data we obtain are mainly real tables after desensitization. Based on the information provided by these tables and manual processing, we construct a database containing 37 data tables. After that, we construct SQL questions based on the database. On this basis, this experiment constructs a Text2SQL question-SQL pair dataset. The dataset was divided into 9792 training set samples and 1088 test set samples at a ratio of 9:1. Each sample contains input questions and corresponding SQL query statements. The specific format of each sample is shown in Figure \ref{fig6}. 

\subsection{Training details}
In the table selection subtask and the column selection subtask, we select the BERT model, which contains 12 encoders, 12 attention heads, and 768 input dimensions. In the SQL generation subtask and the value filling subtask, we select the T5 model, which contains 12 encoders and decoders, 12 attention heads, and 768 input dimensions.

\begin{table}[ht]
\renewcommand{\arraystretch}{1.3}
\caption{comparison experiment}
\label{table1}
\centering
\begin{tabular}{cc}
  \toprule
		Model & Logic Form Accuracy   \\
  \midrule
		IRNET \cite{guo2019towards}& 36.4\%  \\
		IGSQL  \cite{cai2020igsql}& 68.0\%  \\
        RAT-SQL \cite{wang2019rat}& 81.5\%  \\
        SPSQL (Our model) & \textbf{95.6\%} \\
  \bottomrule
\end{tabular}
\end{table}

\begin{table*}
\renewcommand{\arraystretch}{1.3}
\caption{Ablation Experiment and Data augmentation Experiment}
\label{table2}
\centering
\begin{tabular}{ccccc}
  \toprule
		Model & 
  \makecell{Initial \\ Dataset} & 
  \makecell{Expanded By \\ Synonymous Substitution} & \makecell{Add Or Delete \\ Columns} &
  \makecell{Expanded By \\ simBERT}  
  \\
  \midrule
		T5+NER & 83.4\% & 89.3\% & 91.0\% & 93.8\%\\
		T5+T5 & 81.2\% & 87.6\% & 89.7\% & 92.0\%\\
        T5+T5+NER (Our model)& \textbf{85.0\%} & \textbf{92.5\%} & \textbf{93.3\%} & \textbf{95.6\%}\\
  \bottomrule
\end{tabular}
\end{table*}
This paper sets up three types of experiments: comparison experiment, ablation experiment, and data augmentation experiment. Experiments use Logic Form Accuracy to evaluate the accuracy of SQL generation. In the comparison experiment, to study the effect of our model and other end-to-end models, we experiment on our model (SPSQL), IGSQL \cite{cai2020igsql}, RAT-SQL \cite{wang2019rat}, and IRNET \cite{guo2019towards}. In the ablation experiment, we merge the SQL generation subtask and the value filling subtask and implement them with a T5 model to study the impact of different structures in the SQL generation; We remove the NER module in our method to study the impact of the NER module. In the data augmentation experiment, to compare the effects of different data augmentation methods, we set up four datasets: initial dataset, dataset expanded by synonymous substitution, dataset expanded by random addition and deletion columns, and dataset expanded by simBERT. The initial dataset only contains the data augmentation method of replacing the question keywords. To make the database more relevant to the user's common expressions, we generate new question-SQL pairs to the dataset by replacing partial expressions with synonyms. In case the sequence labeling model sometimes does not select some columns, we randomly add and delete columns to generate new question-SQL pairs to the dataset. To diversify the expression of questions, we use simBERT to generate questions that are more than 95\% similar to the original questions and add them to the dataset. On the one hand, this increases the diversity of question-SQL pairs, on the other hand, it expands the dataset. The dataset expanded by simBERT is 5 times the initial dataset, and the comparison experiment is carried out under the dataset expanded by simBERT.

\subsection{Results}
The result of the comparison experiment is shown in Table \ref{table1}. The logic form accuracy of different models varies greatly. The lowest of the three end-to-end methods is IRNET, and the highest is RAT-SQL. Our method SPSQL is higher than RAT-SQL, reaching 95.6\%, an increase of 17.3\%, which fully demonstrates the superiority of SPSQL to practical application scenarios with small data volume. The result of the ablation experiment is shown in Table \ref{table2}. No matter in which dataset, the accuracy of our method is higher than that of “T5+NER” and “T5+T5”; “T5+T5+NER” is 2.49\% higher than “T5+NER” on average, which shows that splitting the SQL generation model into SQL generation submodel and value filling submodel can improve the accuracy of SQL statement generation. The result of the data augmentation is shown in Table \ref{table2}. The accuracy of directly using the initial dataset is the lowest. After the expansion of the synonymous substitution question-SQL pairs, the average increase is 7.93\%. After the random addition and deletion of the column information, the average increase is 9.78\%. However, their effect is worse than the simBERT method, which is 12.75\% higher than the accuracy of the initial dataset.

\section{Conclusion}
In this work, we propose a text2SQL model SPSQL. It is constructed into a pipe structure by table selection model, column selection model, SQL generation model, and value filling model. It receives the text input by the user and generates the corresponding standard SQL query statement through the table selection task, column selection task, SQL generation task, and value filling task in turn. We first explore the SQL statement generation effect of different methods. The experimental results show that, compared with other methods, the logic form accuracy of SQL statements generated by SPSQL is the highest, reaching 95.6\%, in the face of practical application scenarios with small data volume. Then we explore the data augmentation method, and the dataset augmented by our method increases the accuracy of SPSQL by 12.5\%. We also construct the NER module to solve the problem of incorrect SQL generation caused by inaccurate entity recognition due to insufficient training data. Its introduction increases the accuracy of SPSQL by 3.6\%. In future applications, Text2SQL will develop toward solving complex query requests and further improving automation and accuracy.


\section*{Acknowledgment}

This work is supported by Zhejiang Electric Power Co., Ltd. Science and Technology Project (No. 5211YF220006).



%
\bibliographystyle{IEEEtran} 
\bibliography{1} 

\end{CJK*}
\end{CJK*}
\end{document}